\documentclass[10pt,twocolumn,letterpaper]{article}

\usepackage{iccv}
\usepackage{times}
\usepackage{epsfig}
\usepackage{graphicx}
\usepackage{amsmath}
\usepackage{amssymb}
\usepackage{url} 
\usepackage{{subfig}}
\usepackage{enumitem}
\usepackage{mathtools}
\usepackage{lipsum}
\usepackage{adjustbox}
\usepackage[normalem]{ulem}
\usepackage[toc,page]{appendix}
\allowdisplaybreaks
\usepackage[pagebackref=true,breaklinks=true,letterpaper=true,colorlinks,bookmarks=false]{hyperref}

\iccvfinalcopy 

\def\iccvPaperID{3323} 

\ificcvfinal\fi
\begin{document}

\title{Stochastic Dynamics for Video Infilling}
\author{\hspace*{-10pt}Qiangeng Xu $^{,1}$ \qquad Hanwang Zhang$^{,2}$ \qquad Weiyue Wang$^1$
	\qquad Peter N. Belhumeur$^3$  \qquad Ulrich Neumann$^1$\\
	\hspace{-10mm}$^1$University of Southern California\hspace{10mm} $^2$Nanyang Technological University
	\hspace{10mm}$^3$Columbia University\\
	{\tt\small \hspace{0mm}\{qiangenx,weiyuewa,uneumann\}@usc.edu}\hspace{5mm}{\tt\small hanwangzhang@ntu.edu.sg}\hspace{5mm}{\tt\small belhumeur@cs.columbia.edu}\qquad
}

\maketitle

\begin{abstract}
    In this paper, we introduce a stochastic dynamics video infilling (SDVI) framework to generate frames between long intervals in a video. Our task differs from video interpolation which aims to produce transitional frames for a short interval between every two frames and increase the temporal resolution. Our task, namely video infilling, however, aims to infill long intervals with plausible frame sequences. Our framework models the infilling as a constrained stochastic generation process and sequentially samples dynamics from the inferred distribution. SDVI consists of two parts: (1) a bi-directional constraint propagation module to guarantee the spatial-temporal coherence among frames, (2) a stochastic sampling process to generate dynamics from the inferred distributions. Experimental results show that SDVI can generate clear frame sequences with varying contents. Moreover, motions in the generated sequence are realistic and able to transfer smoothly from the given start frame to the terminal frame. Our project site is \url{https://xharlie.github.io/projects/project_sites/SDVI/video_results.html}
\end{abstract}

\section{Introduction}
    Video temporal enhancement is generally achieved by synthesizing frames between every two consecutive frames in a video. Recently, most studies \cite{meyer2018phasenet,jiang2017super} focus on interpolating videos with frame rate above 20 fps.  The between-frame intervals of these videos are short-term and the consecutive frames only have limited variations.
    Instead, our study focuses on the \textbf{long-term interval infilling} for videos with frame rate under 2 fps. This study can be applied on recovering low frame rate videos recorded by any camera with limited memory, storage, network bandwidth or low power supply (e.g., outdoor surveillance devices and webcam with an unstable network). 
    
\begin{figure}[!t]
    \centering
      \includegraphics[width=\linewidth]{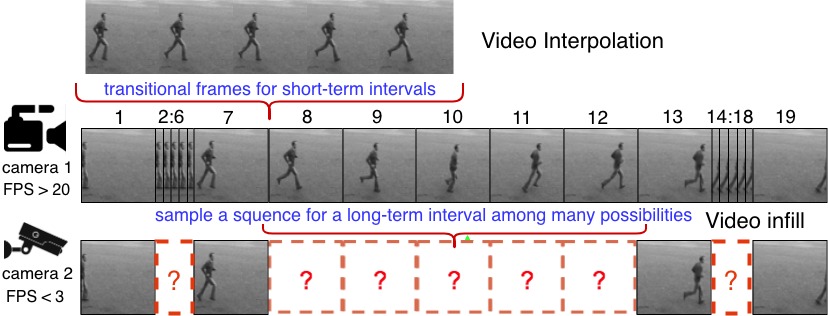}
      \caption{Difference between video interpolation and video infilling. Camera 1 captures frames 1 to 19. Video interpolation aims to generate 5 frames between frame 7 and 8. A low frame rate camera 2 only captures frame 1, 7, 13 and 19. Video infilling focuses on generating a plausible intermediate dynamic sequence for camera 2 (a plausible sequence can be different from the frames 8 to 12).}
      \label{fig:problem_formulation}
    \end{figure}

    The difference between video interpolation and video infilling is shown in Figure \ref{fig:problem_formulation}. Conditioned on frame 7 and 8, video interpolation generates transitional frames containing similar content for short intervals. However, video infilling generates frames in a long-term interval (from frame 8 to 12) and requires the model to produce varying content. At each timestamp, the model needs to sample a plausible dynamic sequence out of many possible movements.  
    
\begin{figure}[!t]
    \centering
      \includegraphics[width=\linewidth]{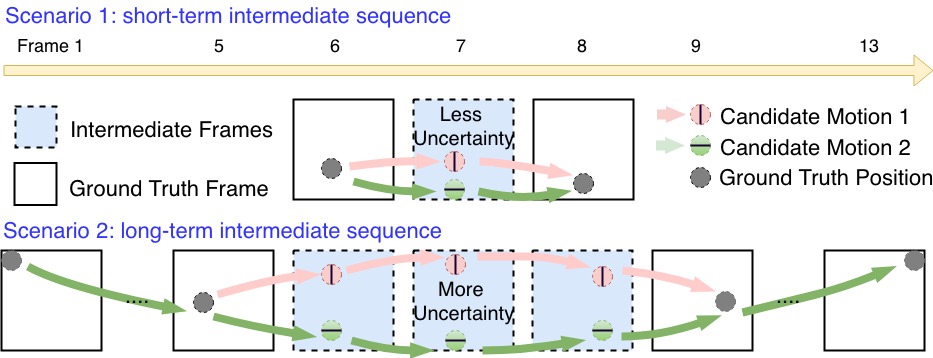}
      \caption{The difference of the randomness between short-term and long-term intervals: The camera in scenario 1 can capture every other frame and the camera in scenario 2 captures 1 frame for every 4 frames. The red and the green trajectories indicate two possible motions in each scenario.}
      \label{fig:stochasticity}
    \end{figure}

    Figure \ref{fig:stochasticity} illustrates the stochastic nature of the long-term intermediate sequence. We observe the following two phenomena: (1) Compared with Scenario 1, since both the interval length and the difference between the two reference frames are larger, the uncertainties in the long-term interval (Scenario 2) are greater. (2) Taken frame 5 and 9 as references, both the red and the green motions between frame 5 and 9 are plausible. If we also add frame 1 and 13 as references, only the green motion is plausible. Consequently, utilizing long-term information (frame 1 and 13 in Figure \ref{fig:stochasticity}) can benefit the dynamics inference and eliminate the uncertainties. Given start and end frames of a long-term interval in a video, we introduce stochastic dynamic video infilling (SDVI) framework to generate intermediate frames which contain varying content and transform smoothly from the start frame to the end frame.
        
\subsection{Task Formulation}
     Following the standard input setting of temporal super-resolution, we formulate our task as follows: For a sequence \(\mathbf{X}\),  only one out of every \(u\) frames (\(u = T-S\)) is captured. The goal of SDVI is to infill a sequence \(\tilde{X}_{S+1:T-1}\) between reference frames \(X_S\) and \(X_T\). In Figure \ref{fig:problem_formulation}, \(X_S\) and \(X_T\) are frame 7 and 13. We also use additional frames (frame 1 and 19) as extended references. \(X_S\), \(X_T\) and extended reference frames (here we choose i=1) form the reference set ``window of reference" \(\mathbf{X_{WR}}\). \vspace{-6pt}
    \[
        \underbrace{X_{S-i\times u},...,X_{S-u}}_{\text{extended reference frames}},X_S,\underbrace{X_{S+1:T-1}}_{\text{our target}},X_T,\underbrace{X_{T+u},..., X_{T+i\times u}}_{\text{extended reference frames}}
    \]
    Different from all existing methods, SDVI inference \(P(X_{S+1:T-1}|\mathbf{X_{WR}})\) instead of \(P(X_{S+1:T-1}|X_{S}, X_{T})\). 
    
\subsection{Model Overviews}
     Most video interpolation methods \cite{niklaus2018context,liu2017video} rely on estimating the short-term pixel movements.
     Our task is also related to video prediction. Video prediction models \cite{wichers2018hierarchical,mathieu2015deep,finn2016unsupervised} can generate long-term sequences by explicit dynamics modeling, but do not take discontinuous inputs and are not optimized for bi-directional constraints. On the contrary, our model explicitly inference the motion dynamics of the interval and propagate constraints from both sides (start and end frames).
     
     Different from both video interpolation and video prediction, the task has three major challenges:
     
    \textbf{1}. The inputs of the video prediction are consecutive frames so the initial momentum is given. However, the inputs of video infilling are sparse and discontinuous ($X_S$ and $X_T$), which makes the task more challenging. 
        
    \textbf{2}. The observation of the last frame becomes a long-term coherence requirement, which gives more constraints to our model. Video prediction only needs to generate visually plausible frames smoothly transferred from previous frames, while video infilling is also required to guarantee the coherence between the previous sequence ($\tilde{X}_{S+1:T-1}$) and the terminal frame $X_{T}$.
    
    \textbf{3}. As illustrated in Figure \ref{fig:stochasticity}, compare with interpolation models, an interval our model needs to infill has more uncertainties, even with more reference frames (frame 1 and 13). 
        
    To inference the initial and final momentum, we expose extended reference frames both from the past (frame 1 and 7 in Figure \ref{fig:problem_formulation}) and the future (frame 13 and 19) to the model. To achieve long-term coherence, we introduce \({RBConvLSTM}\), a multi-layer bi-directional ConvLSTM with residual connections between adjacent layers. The dynamics from both sides are gradually propagated to the middle steps and create dynamic constraint vectors to guide the inference step by step.
    
    To model the uncertainty in the interval, we propose a stochastic model under the bi-directional constraints. At step t, a distribution for an embedding vector is inferred, conditioned on previously generated frames and the reference frames. We sample an embedding vector from the distribution and use a decoder to generate the frame at step t. 
    
    We design our objective function by optimizing a variational lower bound (see \ref{seg:loss}). SDVI achieves state-of-the-art performance on 4 datasets. We also infill every between-frame interval of a real-world video (2fps) and connect them to create an enhanced long video of 16fps (See the video web page in the supplemental material). To summarize, our contributions are:
    \begin{itemize}[noitemsep, topsep=2pt]
    \item To the best of our knowledge, it is the first stochastic model and the first study utilizes the extended frames away from the interval to solve the video infilling.
    \item A module \(RBConvLSTM\)(see \ref{seg:RBConvLSTM}) is introduced to enforce spatial-temporal coherence.
    \item A spatial feature map is applied in the sampling to enable spatial independence of different regions. 
    \item A metric LMS (see \ref{sec:Experiments}) is proposed to evaluate the sequence temporal coherence. 
    \end{itemize}

\section{Related Works}
    Most studies of video interpolation \cite{ilg2017flownet,jiang2017super} focus on generating high-quality intermediate frames in a short-term interval. Since we focus on long-term sequence infilling, our framework adopts long-term dynamics modeling. Therefore we also refer to the studies of video prediction which have explored this area from various perspectives. 
    \subsection{Video Interpolation}
        Video interpolation generally has three approaches: optical flow based interpolation, phase-based interpolation, and pixels motion transformation. Optical flow based methods \cite{herbst2009occlusion,yu2013multi,ilg2017flownet} require an accurate optical flow inference. However, the optical flow estimation is known to be inaccurate for a long time interval. Estimating motion dynamics becomes a more favorable option. The phase-based methods such as \cite{meyer2015phase} modify the pixel phase to generate intermediate frames. Although the strategy of propagating phase information is elegant, the high-frequency and drastic changes cannot be properly handled. The inter-frame change will be more significant in our long-term setting. Currently studies \cite{liu2017video,DBLP:journals/corr/abs-1708-01692,jiang2017super} use deep learning methods to infer the motion flows between the two frames. By far, this branch of approaches achieves the best result and has the potential to solve our task. In our evaluation, we use SepConv \cite{niklaus2017video} and SuperSloMo \cite{jiang2017super} as comparisons.
    \subsection{Deterministic Video Prediction}
        The mainstream video prediction methods take short consecutive sequences as input and generate deterministic futures by iteratively predicting next frame. \cite{vondrick2016generating,mathieu2015deep} use a convolutional network to generate each pixel of the new frame directly. Studies such as \cite{srivastava2015unsupervised,finn2016unsupervised} use a recurrent network to model the dynamics and improve the result drastically. \cite{xingjian2015convolutional} introduces ConvLSTM, which has been proved to be powerful in spatial sequence modeling. \cite{villegas2017decomposing,denton2017unsupervised} propose to model the content and the dynamics independently to reduce the workload for the networks. \cite{tulyakov2017mocogan,villegas2017decomposing} incorporate GANs \cite{goodfellow2014generative} into their model and improve the quality. Notably, two of the generative models \cite{lu2017flexible} and \cite{cai2017deep} can also conduct video completion. However both methods, due to their forward generation mechanism, cannot hold the coherence between the last frame and the generated sequence. SDVI adopts the decomposition of the motion and the content, uses the ConvLSTM in the motion inference and iteratively generates the frame. However, we do not use GANs since our study focuses more on dynamics generation. We also compare SDVI with FSTN in \cite{lu2017flexible}, a prediction model that can also handle video infilling.
    \subsection{Stochastic Video Generation}
        After \cite{babaeizadeh2017stochastic,henaff2017prediction,walker2016uncertain} shows the importance of the stochasticity in video prediction, later studies such as \cite{lee2018stochastic,denton2018stochastic} also conduct the prediction in the form of stochastic sampling. The stochastic prediction process consists of a deterministic distribution inference and a dynamic vector sampling. We also adopt this general procedure. Since SVG-LP introduced in (\cite{denton2018stochastic}) is one of the state-of-the-art models and very related to our study, we use the SVG-LP to compare with SDVI. A concurrent work \cite{jayaraman2018time} can generate an intermediate frame between two given frames. However, the model inclines to generate the frame at the time with low-uncertainty. Therefore their model cannot solve the infilling task since the generated sequence does not have a constant frame density. 
        
\section{Model Details}
    \begin{figure}[!t]
    \centering
      \includegraphics[width=\linewidth]{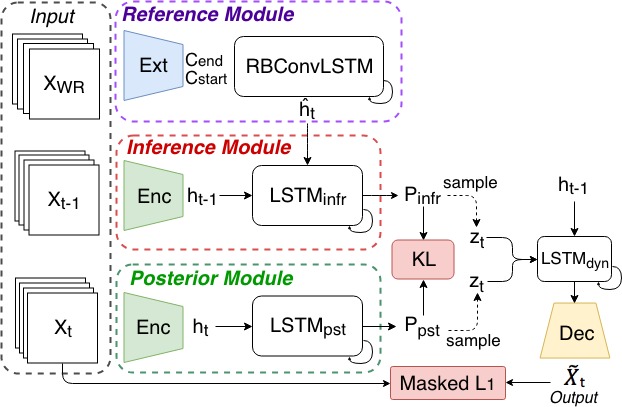}
      \caption{Training of SDVI: All \(Encoder\) (green) share the same weights. The blue and the yellow network are \(Extractor\) and \(Decoder\). Reference module creates dynamic constraint \(\hat{h}_t\) at each step. At step \(t\), Inference module takes \(X_{t-1}\) and \(\hat{h}_t\), while Posterior module takes \(X_{t}\). Inference module and Posterior module will produce different \(z_t\) and therefore different output frames \(\tilde{X}^{infr}_t\) and \(\tilde{X}^{pst}_t\).}
      \label{fig:stochastic_training}
    \end{figure}
    As illustrated in Figure \ref{fig:stochastic_training}, SDVI consists of 3 major modules: Reference, Inference and Posterior modules. Given reference frames \(\mathbf{X_{WR}}\), Reference module propagates the constraints and generate a constraint vector $\hat{h}_t$ for time t. Inference module takes $\hat{h}_t$ and inference an embedding distribution $P_{infr}$ based on $X_{S:t-1}$. Posterior module inference another embedding distribution $P_{pst}$ based on $X_{S:t}$. We sample an embedding vector $z_t$ from $P_{infr}$ and another $z_t$ from $P_{pst}$. A decoder is used to generate a frame \(\tilde{X}_t\) for a given $z_t$. During training, we use KL divergence to minimize the distance between $P_{infr}$ and $P_{pst}$. At test, Posterior module is not required and $z_t$ is sampled from $P_{infr}$. 
    We list the notations as follows:
    \begin{itemize}[noitemsep, leftmargin=10pt]
    \item[*] t: a time step between start step S and terminal step T. 
    \item[*] S:t: the sequence start from step S to step t.
    \item[*] \(X_t\): The ground truth frame at time step \(t\).
    \item[*] \(\tilde{X}_t\): The frame generated on step t. 
    \item[*] \(C_{start}\) and \(C_{end}\): Momentum vectors extracted from \(\mathbf{X_{WR}}\), used as initial cell states for \(RBConvLSTM\).
    \item[*] \(h_t\): The dynamic vector extracted from $X_t$. 
    \item[*] \(\hat{h}_t\): The constraint vector at the step \(t\).
    \item[*] \(P_{infr}\) and \( P_{pst}\): The distributions of the embedding vector generated by Inference and Posterior module.
    \item[*] \(z_t\): The embedding vector on time step \(t\). \(z_t^{infr}\) is sampled from \(P_{infr}\) and \(z_t^{pst}\) is sampled from \( P_{pst}\). 
    \end{itemize}
    
    \subsection{Reference Module}
        Reference module includes an \(Extractor\) and a \(RBConvLSTM\).  Given all the frames in \(\mathbf{X_{WR}}\), the \(Extractor\) learns the momentum and output two vectors \(C_{start}\) and \(C_{end}\). With the dynamics and momentum of $X_S$ and $X_T$, \(RBConvLSTM\) outputs a constraint vector \(\hat{h}_t\) for each intermediate step \(t\). The whole sequence of the constraint vector has a conditional distribution \(P(\hat{h}_{S:T}|\mathbf{X_{WR}})\). 
        \paragraph{RBConvLSTM}
            \label{seg:RBConvLSTM}
            \begin{figure}[!t]
              \includegraphics[width=\linewidth]{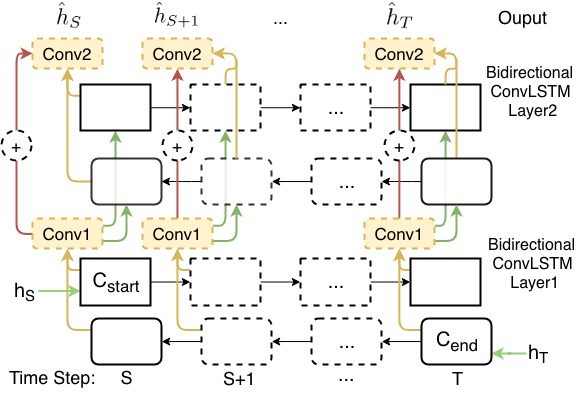}
              \caption{A two layers RBConvLSTM: The initial cell states of the first layer are assigned as \(C_{start}\) and \(C_{end}\). \(h_S\) and \(h_T\) are taken as inputs. Combined with the residuals (red arrows), each layer's outputs (yellow arrows) would go through a convolution module and become the inputs (green arrows) to the next layer.}
              \label{fig:stochastic_RBConvLSTM}
            \end{figure}
            \(RBConvLSTM\), a residual bi-directional ConvLSTM, is based on the studies of seq2seq \cite{sutskever2014sequence,bahdanau2014neural,xingjian2015convolutional}. As shown in Figure \ref{fig:stochastic_RBConvLSTM}, the first layer of \(RBConvLSTM\) uses \(C_{start}\) as the initial state of the forward cell and \(C_{end}\) for the backward cell. We need to propagate sparse constraints \(h_S, h_T\) to every time step from $S+1$ to $T-1$ to get outputs \(\hat{h}_S, \hat{h}_{S+1},...,\hat{h}_T\) as constraint vectors for Inference Module. They are critical to achieve the long-term coherence. Since the input features to the bottom layer \(h_S, \mathbf{0},...,h_T\) share the same feature space with \(\hat{h}_{S:T}\), inspired by \cite{he2015deep}, we add an residual connection between each two layers to elevate the bottom features directly to the top. In the end, RBConvLSTM combines all the three structures: the ConvLSTM, the bi-directional RNN and the residual connections.
    \subsection{Inference Module}
         As shown in Figure \ref{fig:stochastic_RBConvLSTM}, We extract a dynamic vector \(h_t-1\) from each \(X_t-1\). \(LSTM_{infr}\) takes the \(h_{t-1}\) and the constraint vector \(\hat{h_t}\), then infers a distribution \(P_{infr}\) of a possible dynamic change. This module resembles the prior distribution learning of stochastic prediction, however, \(P_{infr}\) here is written as \(P_{infr} = P(z_t|X_{S:t-1}, \mathbf{X_{WR}})\). 
        
    \subsection{Posterior Module} 
         A generated sequence \(\tilde{X}_{S+1:T-1}\) can still be valid even it is different from the ground truth \(X_{S+1:T-1}\). Therefore our model need to acquire a target distribution \(P_{pst}\) for step t, so Inference module can be trained by matching \(P_{infr}\) to the target. Here we expose the frame \(X_t\) to Posterior module, so it can generate a posterior distribution \(P_{pst} = P(z_t|X_{S:t})\) for \(P_{infr}\) to match. 
         
    \subsection{Training and Inference}
        From \(P_{pst}\), we can sample a embedding vector vector \(z_t^{pst}\). Conditioned on the previous ground truth frames and \(z_t^{pst}\), the \(Decoder\) generates the \(\tilde{X}_t^{pst}\). Separately, we also sample a vector \(z_t^{infr}\) from \(P_{infr}\), and generate the \(\tilde{X}_t^{infr}\) in the same way. 
        
\begin{figure}
        \centering
          \includegraphics[width=\linewidth]{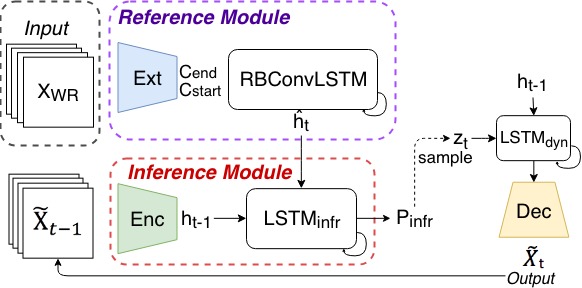}
          \caption{Inference of SDVI: Without ground truth frame \(X_{t-1}\), the generated frame \(\tilde{X}_{t-1}\) serves as the input to Inference module on step \(t\).}
          \label{fig:stochastic_generation}
        \end{figure}

        Since the ground truth frames \(X_t\) is not available at time t, we can only use Inference module to sample a \(z_t\). The inference pipeline is shown in Figure \ref{fig:stochastic_generation}.

    \subsection{Dynamic Spatial Sampling}\label{Sampling}
        \begin{figure}[!t]
          \includegraphics[width=1\linewidth]{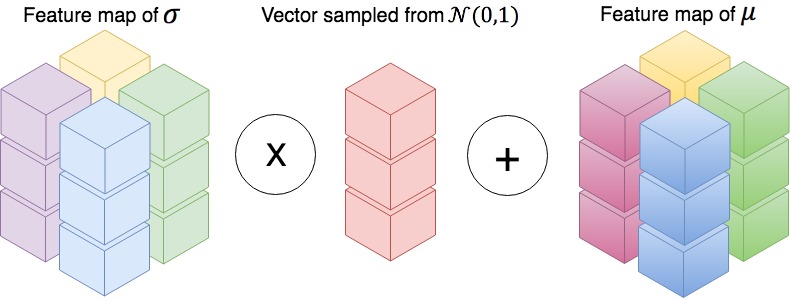}
          \centering
          \caption{The sampled vector (in the middle) is applied on all locations.}
          \label{fig:stochastic_spatial_sampling}
        \end{figure}
         Using the re-parameterization trick \cite{kingma2013auto}, we model \(P_{pst}\) and \(P_{infr}\) as Gaussian distributions \(N_{pst}(\mu_t,\sigma_t)\) and \(N_{infr}(\mu_t,\sigma_t)\). Different locations in one frame may have different levels of uncertainty. Uniformly draw a sample following the same distribution everywhere will hinder the modeling (see SDVI non-spatial in Table \ref{tab:avg_overall}). Consequently, we introduce a spatial sampling process (Figure \ref{fig:stochastic_spatial_sampling}). Instead of using vectors \cite{denton2018stochastic}, we use spatial feature maps for \(\mu_t\) and \(\sigma_t\). To get the \(z_t\), we multiply the sampled vector on each location of \(\sigma_t\), then add the \(\mu_t\) on the product.
    \subsection{Loss Function}
    \label{seg:loss}
        \paragraph{Pixel Loss}
            To make the \(\tilde{X}_t^{pst}\) reconstruct real \(X_t\), we introduce a pixel reconstruction loss \(L_1(X_t, \tilde{X}_t^{pst})\).  We also observe that imposing a pixel prediction loss to the \(\tilde{X}_t^{infr}\) after the \(P_{infr}\) getting stable can further improve the video quality during inference. 
        \paragraph{KL Divergence Loss}
            \(P_{pst}(z_t|X_{S:t})\) carries the dynamics from the \(h_t\) to reconstruct the \(X_t\). Since we use only Inference module during inference, the \(P_{infr}(z_t|X_{S:t-1},  \mathbf{X_{WR}})\) needs to predict the embedding vector alone. Therefore, we also add two KL divergences between \(P_{infr}\) and \(P_{pst}\):
            {\small
                \begin{align}
                        L_{KL} = D_{KL}(P_{pst}||P_{infr}) + D_{KL}(P_{infr}||P_{pst})
                \end{align}
            }
            Both the forward and the reverse KL-divergence of \(P_{infr}\) and \(P_{pst}\) achieve the minimum when the two distributions are equal. However, according to \cite{fox2012tutorial}, since \(D_{KL}(P_{pst}||P_{infr})\) is sampled on \(P_{pst}\), it will penalize more when \(P_{pst}\) is large and \(P_{infr}\) is small. Therefore this term will lead to a \(P_{infr}\) with higher diversity. On the other hand, \(D_{KL}(P_{infr}||P_{pst})\) will make the inference more accurate when \(P_{infr}\) has large value. To better serve our task, we decide to keep both terms to strike a balance between accuracy and diversity.
        \paragraph{Full Loss}
            Overall, our final objective is to minimize the combined full loss:
            {\small
                \begin{align}
                    & L_C = \sum_{t=S+1}^{T-1} [\underset{\text{pixel reconstruction loss}}{\beta \cdot L_1(X_t, \tilde{X}_t^{pst})} + \underset{\text{pixel prediction loss}}{(1-\beta) \cdot L_1(X_t, \tilde{X}_t^{infr})} \nonumber  \\
                    &+ \underset{\text{inclusive KL loss}}{\alpha \cdot D_{KL}(P_{pst}||P_{infr})}  + \underset{\text{exclusive KL loss}}{\alpha \cdot D_{KL}(P_{infr}||P_{pst})}]  \label{obj}
                \end{align}
            }
            The \(\beta\) balances the posterior reconstruction and the inference reconstruction, while the \(\alpha\) determines the trade-off between the reconstruction and the similarity of the two distributions. To show the effectiveness of these loss terms, we also compare the full loss \eqref{obj} with a loss only composed of the pixel reconstruction loss and the inclusive KL loss (similar to the loss in \cite{denton2018stochastic}), shown as \text{``SDVI loss term 1\&3"} in Table \ref{tab:avg_overall}. We also provide the theoretical explanation of the full loss in the supplemental material.

\section{Experiments}
    \label{sec:Experiments}
    \paragraph{Datasets} We first test SDVI on 3 datasets with stochastic dynamics: Stochastic Moving MNIST(SM-MNIST) \cite{denton2018stochastic} with random momentum after a bounce, KTH Action Database \cite{schuldt2004recognizing} for deformable objects and BAIR robot pushing dataset \cite{ebert2017self} for sudden stochastic movements. We also compare with the interpolation models on a challenging real-world dataset, UCF101\cite{soomro2012ucf101}. 
    
    \paragraph{Last Momentum  Similarity and Other Metrics}
    Three metrics are used for quantitative evaluation: structural similarity (SSIM), Peak Signal-to-Noise Ratio (PSNR), and Last Momentum  Similarity (LMS). 
    
    An infilled sequence that is different from the ground truth (low SSIM and PSNR) is still valid if it can guarantee the long-term coherence between \(\tilde{X}_{S:T-1}\) and \(X_{T}\). Thus we introduce the last momentum similarity (LMS) calculated by the mean square distance between the optical flow from \(X_{T-1}\) to \(X_{T}\) and the optical flow from \(\tilde{X}^{infr}_{T-1}\) to \(X_{T}\). We find LMS a good indicator of the video coherence since no matter how the dynamic being sampled, both the object's position and speed should make a smooth transition to \(X_T\). 
    
    \paragraph{Movement Weight Map}
    \label{sec:movement_weight_map}
    During training, we apply a movement weight map to each location of the pixel loss to encourage movement generation. For a ground truth frame $X_t$, if a pixel value stays the same in $X_{t-1}$, the weight is 1 on that location. Otherwise, we set the weight to be \(\eta > 1\) to encourage the moving region. This operation helps us to prevent the generation of sequences.
    
    \paragraph{Main Comparison}
    We compare our model with the-state-of-the-art studies of both video interpolation and video prediction (with modification). Except for SuperSloMo, all models are trained from scratch under the same conditions for all datasets.
    
    We select two high-performance interpolation models SepConv\cite{niklaus2017video} and SuperSloMo\cite{jiang2017super}. Due to SepConv's limitation (sequences must have the length of \(2^n - 1\)), all the following evaluations are under the generation of 7 frames. Following their instruction, we complete the training code of SepConv. We can't get the code of SuperSloMo, but we acquire the results from the authors of \cite{jiang2017super}.
    
    Two prediction models are picked: FSTN\cite{lu2017flexible}, a deterministic generation model; SVG-LP\cite{denton2018stochastic}, an advanced stochastic prediction model. Since FSTN and SVG-LP are not designed to solve the infilling task, we concatenate the representation of the last frame \(X_T\) to their dynamic feature maps in each step. Then the SVG-LP simulates SDVI without Reference module, and the FSTN is equivalent to SDVI without Reference and Posterior module. 
    
    \paragraph{Ablation Studies}
    Ablation studies are as follows: 
    
    1. To show that the spatial sampling enables spatial independence, we replace the feature map by a vector in the dynamic sampling process and denote it by ``SDVI non-spatial''. If we up-sample a vector, the information from one area would have an equivalent influence to another area. Therefore it tends to generate a sequence with a single movement (Figure \ref{fig:stochastic_SMMNIST} non-spatial). 
    
    2. To show the benefit of extra reference frames, we remove the extended reference frames in \(\mathbf{X_{WR}}\). We denote this setting by ``SDVI without 0 \& 24''.
    
    3. Our loss has two more terms than another stochastic model \cite{denton2018stochastic}. Therefore we also conduct experiments with only the pixel reconstruction loss and the inclusive KL loss. We denote this setting by ``SDVI loss term 1 \& 3''. \\
    
    Since our model is stochastic, we draw 100 samples for each interval as in \cite{babaeizadeh2017stochastic, denton2017unsupervised, lee2018stochastic} and report a sample with the best SSIM.    
    More video results for various settings (see the video web page), dataset details (Appendix D), network architectures and the training details (see Appendix C) can be found in the supplemental material. 
    \begin{table*}[!t]
    \begin{adjustbox}{width=1\textwidth}
    \begin{tabular}{llll|lll|lll|lll}
    
                         &  \multicolumn{3}{c}{SM-MNIST} &  \multicolumn{3}{c}{BAIR}  &    \multicolumn{3}{c}{KTH}   &             \multicolumn{3}{c}{UCF101} \\
                         & PSNR            & SSIM           & LMS            & PSNR            & SSIM           & LMS           & PSNR            & SSIM           & LMS            & PSNR            & SSIM           & LMS             \\
    SDVI full model       & \textbf{16.025} & \textbf{0.842} & \textbf{0.503} & 21.432          & \textbf{0.880} & \textbf{1.05} & 29.190          & 0.901          & \textbf{0.248} & \textbf{16.635} & \textbf{0.598} & \textbf{15.678} \\
    SDVI without 0 \& 24 & 14.857          & 0.782          & 1.353          & 19.694          & 0.852          & 1.360          & 26.907          & 0.831          & 0.478          & ---             & ---            & ---             \\
    SDVI non-spatial     & 13.207          & 0.752          & 6.394          & 19.938          & 0.865          & 1.159         & 29.366          & 0.896          & 0.276          & ---             & ---            & ---             \\
    SDVI loss term 1\&3  & 15.223          & 0.801          & 0.632          & 19.456          & 0.849          & 1.277         & 28.541          & 0.854          & 0.320          & ---             & ---            & ---             \\
    SVG-LP               & 13.543          & 0.741          & 5.393          & 18.648          & 0.846          & 1.891         & 28.131          & 0.883          & 0.539          & ---             & ---            & ---             \\
    FSTN                 & 14.730          & 0.765          & 3.773          & 19.908          & 0.850          & 1.332         & \textbf{29.431} & 0.899          & 0.264          & ---             & ---            & ---             \\
    SepConv              & 14.759          & 0.775          & 2.160          & \textbf{21.615} & 0.877          & 1.237         & 29.210          & \textbf{0.904} & 0.261          & 15.588          & 0.443          & 20.054          \\
    SuperSloMo           & 13.387          & 0.749          & 2.980           & ---             & ---            & ---           & 28.756          & 0.893          & 0.270          & 15.657          & 0.471          & 19.757         
    
    \end{tabular}
    \end{adjustbox}
    \caption{Metrics averaging over all 7 intermediate frames. We report the scores of the best-sampled sequences for SDVI.}
    \label{tab:avg_overall}
    \end{table*}
  
    \begin{figure}[!t]
      \includegraphics[width=\linewidth]{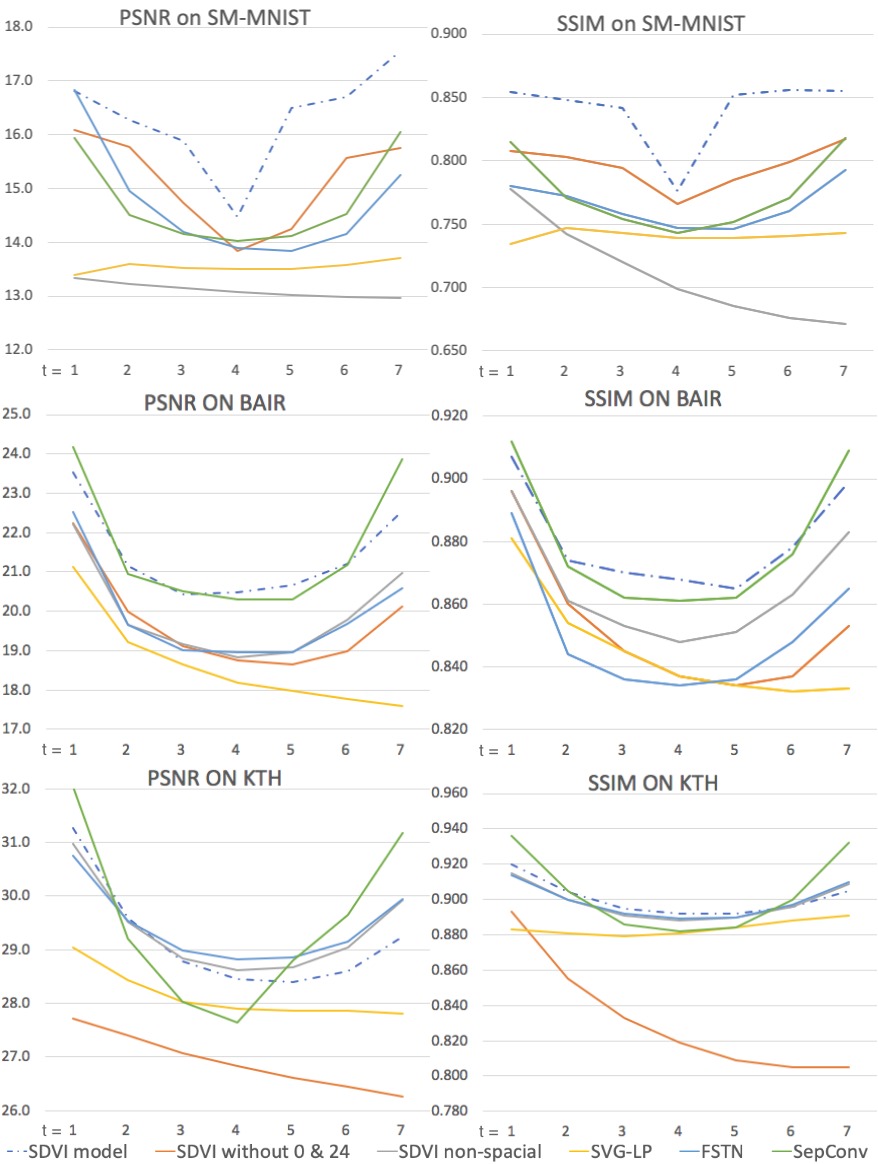}
      \caption{Average PSNR and SSIM at each step in test sets.}
      \label{fig:stochastic-per_frame_evaluation}
    \end{figure}
    \subsection{Stochastic Moving MNIST (SM-MNIST)}
    \begin{figure}[!t]
      \includegraphics[width=\linewidth]{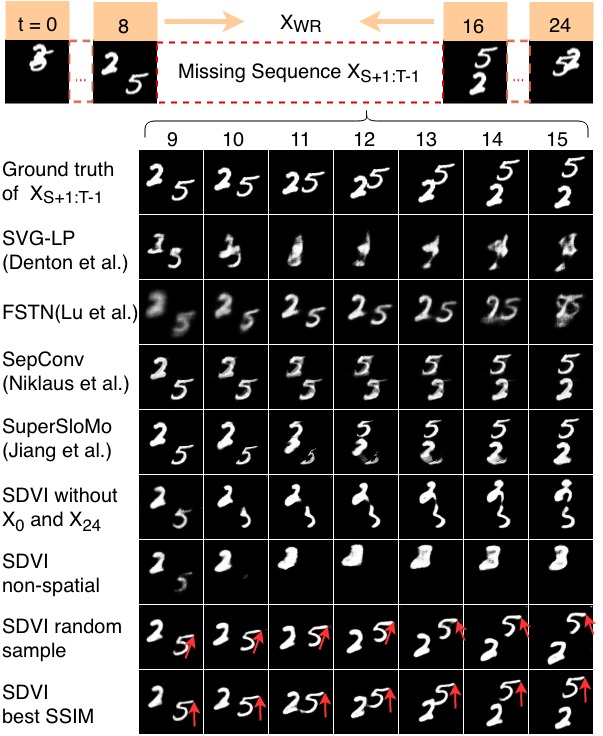}
      \caption{The digit 5 in our best sequence follows the upward trajectory of the ground truth. In another sampled sequence, the 5 goes upper right and then bounce upper left.}
      \label{fig:stochastic_SMMNIST}
    \end{figure}
    
    Digits in SM-MNIST introduced by \cite{denton2018stochastic} will bounce off the wall with a random speed and direction. The uncertainty of the outcome after a bounce makes it a challenging task for all methods. The Avg PSNR, SSIM and LMS over all test frames are shown in Table \ref{tab:avg_overall}. We also plot the metric values averaging on each step in Figure \ref{fig:stochastic-per_frame_evaluation}. Figure \ref{fig:stochastic_SMMNIST} shows the qualitative evaluation for all comparisons. When the two digits in frames 8 and 16 having significant position differences, interpolation models such as SepConv and SuperSloMo would still choose to move the pixel based on the proximity between the two frames: the digits 2 and 5 gradually transfer to each other since the 2 in frame 8 is closer to the 5 in frame 16. Because the deterministic model FSTN cannot handle the uncertainty after a bounce, the model gets confused and generates a blurry result. The SVG-LP cannot converge in this setting since it doesn't have a constraint planning module like the \(RBConvLSTM\) to lead the sequence to the final frame. Without spatial independence, a non-spatial representation cannot sample different dynamics for different areas. The two digits in the result of "SDVI non-spatial" collapse into one, then move toward the final frame. Finally, our full model can learn the bouncing rule and provide plausible alternative sequences. Although our randomly sampled sequence diverges from the ground truth, this sequence can still keep the coherence with frame 8 and 16 under plausible dynamics. 
    \begin{figure}[!t]
      \includegraphics[width=\linewidth]{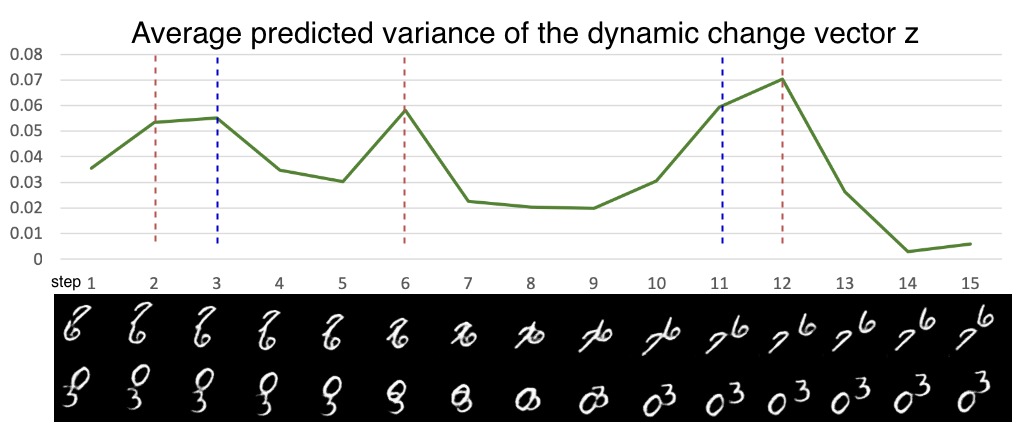}
      \caption{SDVI generates higher variances coincident to the "wall bouncing" event, indicated by the two dash lines(e.g. first sequence: red lines mark the bounces of the digit 6 and blue ones mark the bounces of 7).}
      \label{fig:stochastic-variance}
    \end{figure}
    
    We also study how good our method models the uncertainty as in \cite{denton2018stochastic}. In 768 test sequences, we randomly select two digits for each sequence and synchronize all sequences' trajectories. Figure \ref{fig:stochastic-variance} shows the normalized average variance of the distribution of \(z_t\) for frames 2 to 14 (generated), while frame 1 and 15 are the ground truth frames.
    
    \subsection{BAIR robot pushing dataset}
        \begin{figure}[!htb]
        \centering
          \includegraphics[width=\linewidth]{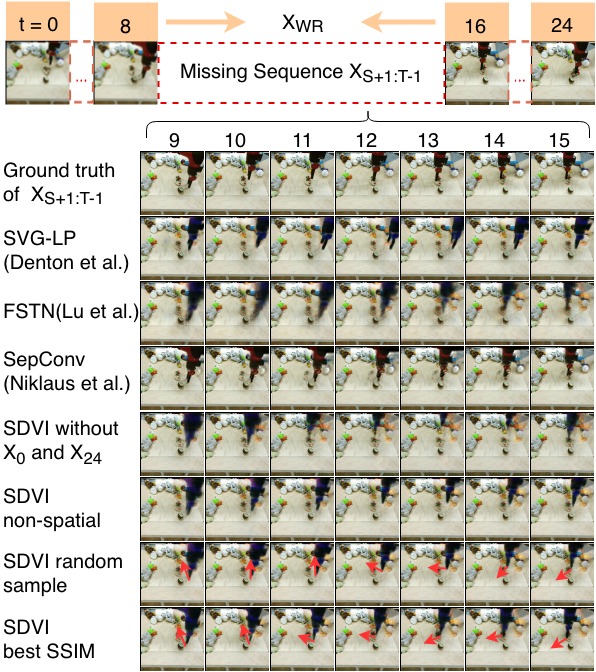}
          \caption{The arm in the best sequence follows the same movements in ground truth: first upward left then downward left. In another sampled sequence, the arm firstly goes straight up and then straight left, finally downward left.}
          \label{fig:stochastic_BAIR}
        \end{figure}
        The BAIR robot pushing dataset \cite{ebert2017self} contains sequences of a robot arm pushing various objects in the RGB domain. The movements of the arm do not follow smooth trajectories, and the movement changes are prompt. As shown in Table \ref{tab:avg_overall}, although our SDVI marginally outperforms other models on SSIM, the SepConv achieves the best PSNR. As shown in Figure \ref{fig:stochastic_BAIR}, since the SepConv relies more on pixel proximity, the shapes of the static objects in this method are nicely preserved. However, SepConv can't model the stochasticity while its movement is simplified to a straight sliding. The frames in the middle suffer the most in all metrics (Figure \ref{fig:stochastic-per_frame_evaluation}). The stochasticity of the movement makes it hard for SVG-LP's arm to go back to the final frame and for FSTN to generate sharp shapes. The objects created by SDVI without spatial sampling are more blurry since all the areas will be disturbed by the change of the dynamics. On the other hand, the result of SDVI without using reference frames 0 and 24 diverges too much away from the ground truth movement. Our full model cannot only sample a similar sequence to the ground truth, but sequences with reasonably varied movements (last two rows in Figure \ref{fig:stochastic_BAIR}).
        \begin{figure*}[!hbt]
        \centering
          \includegraphics[width=\textwidth]{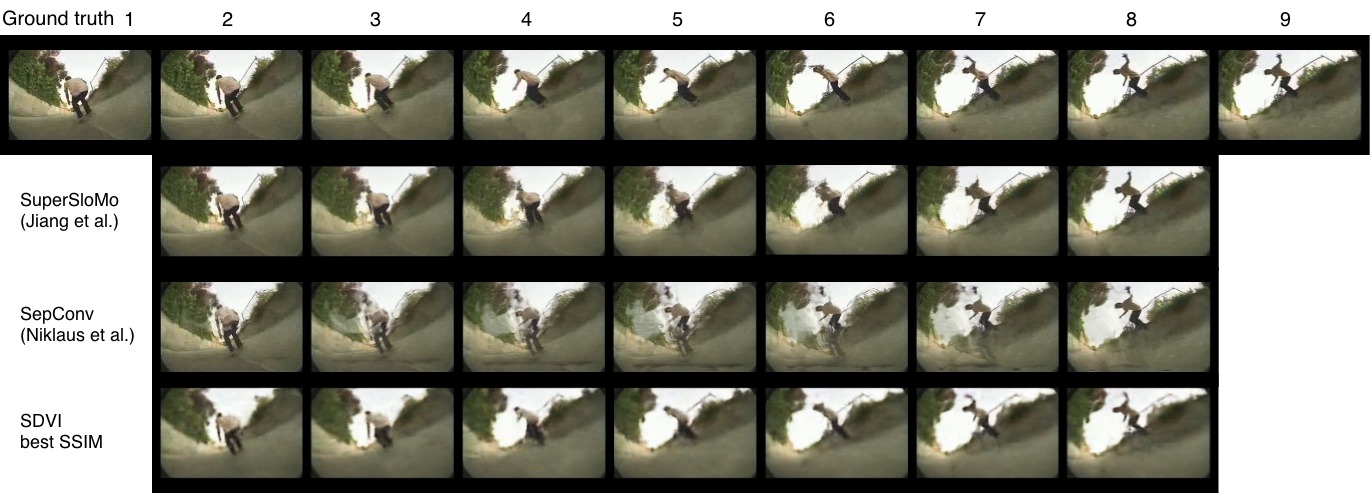}
          \caption{Best view in color. See Appendix E in the supplemental material for more comparisons on UCF101.}
          \label{fig:stochastic_UCF_skateboard}
        \end{figure*}
    \subsection{KTH Action Dataset}
        \begin{figure}[!hbt]
        \centering
          \includegraphics[width=\linewidth]{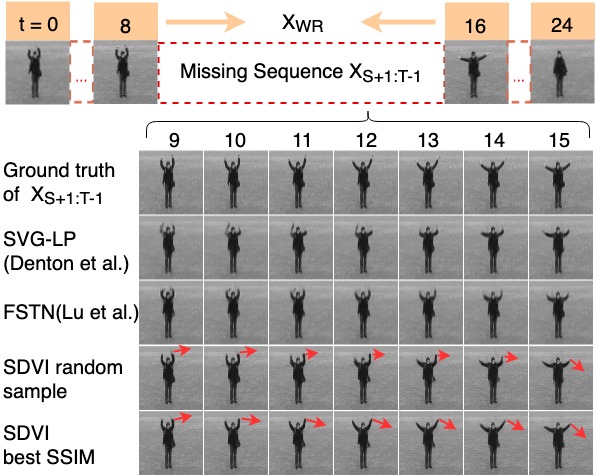}
          \caption{Our best-sampled sequence keeps the arm straight. In a randomly sampled sequence, the forearm bends first then stretches straight in the end.}
          \label{fig:stochastic_KTH2}
        \end{figure}
        \begin{figure}[!hbt]
        \centering
          \includegraphics[width=\linewidth]{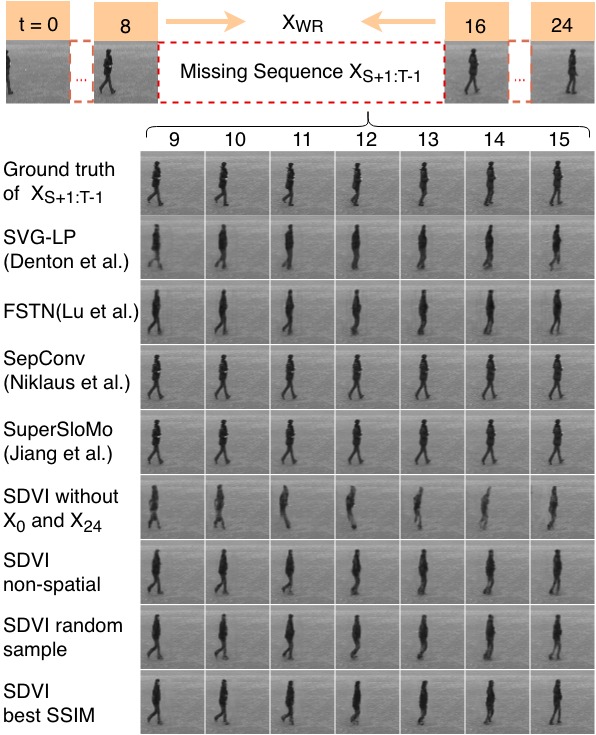}
          \caption{The sliding tendency of SepConv will cause motion errors and high LMS.}
          \label{fig:stochastic_KTH1}
        \end{figure}
        The KTH Action dataset \cite{schuldt2004recognizing} contains real-world videos of multiple human actions. In our setting, all actions are trained together. Although the background is uniform, there is still some pixel noise. We found setting the map's weight to 3 on moving pixels is beneficial. Since most actions such as waving follow fixed patterns, the FSTN and the SepConv can achieve the best scores in PSNR and SSIM (Table \ref{tab:avg_overall}). However, if the object in frame 8 and 16 has a similar body pose, the SepConv and the SuperSloMo will freeze the object's body and slide the object to its new position (Figure \ref{fig:stochastic_KTH1}). SDVI without frame 0 and 24 suffers from the uncertainty of the initial state (Figure \ref{fig:stochastic_KTH1}). The result of FSTN (in \ref{fig:stochastic_KTH2}) contains blurry pixels on moving region although it keeps the static parts sharp. Our sequence with best SSIM has a similar pattern as the ground truth. Even the random sampled sequence shown in Figure \ref{fig:stochastic_KTH2} has different dynamics in the middle, its initial and final movements still stick to the ground truth. Therefore our model still achieves an outstanding LMS over other methods (Table \ref{tab:avg_overall}). 
    \subsection{UCF101}
        Collected from YouTube, UCF101 contains realistic human actions captured under various camera conditions. We train both our model and SepConv on the training set of UCF101, and we get the result from the authors of \cite{jiang2017super}. Our results over 557 test videos outperform both SepConv and SuperSloMo (Table \ref{tab:avg_overall}). In Figure \ref{fig:stochastic_UCF_skateboard}, our model can infer the people's motion consistently. However, it's challenging for SepConv and SuperSloMo to estimate the pixel movement for the middle frames(frame 4 to 6), even though they can generate sharper frames near the two reference frames. 
\section{Conclusions and Future Work}
    We have presented a stochastic generation framework SDVI that can infill long-term video intervals. To the best of our knowledge, it is the first study using extended reference frames and using a stochastic generation model to infill long intervals in videos. Three modules are introduced to sample a plausible sequence that preserves the coherence and the movement variety. Extensive ablation studies and comparisons with the state-of-the-art methods demonstrate our good performance on 4 datasets. A metric LMS is proposed to evaluate the sequence coherence. Although currently SDVI can be iteratively applied to infill an interval with any numbers of frames, its flexibility could be further improved. Another direction is to enhance the generality of the model to work across different video domains.

{\small
\bibliographystyle{ieee}
\bibliography{egbib}
}
\begin{appendices}
\section{Theoretical Explanation of Pixel Prediction Loss and Exclusive KL Loss}
\label{appendix:A}
    We represent the sequence probability of $z_t \sim P_{infr}$ from the start step S to the terminal step T as $P_{infr}(z_{S:T})$.
    
    We represent the sequence probability of $z_t \sim P_{pst}$ from the start step S to the terminal step T as $P_{pst}(z_{S:T})$.
    
    Since we use the output from Inference module as the result during inference phase, we are interested in maximizing a sequence's probability \(P(X_{S:T})\) from the Posterior Module. The logarithm of this probability has the following lower bound:
    {\small
        \begin{align}
            & log P(X_{S:T})  \nonumber \\
            &= log \int_{z_{S:T}} P(X_{S+1:T}|z_{S:T})P_{pst}(z_{S:T})\frac{P_{infr}(z_{S:T})}{P_{infr}(z_{S:T})} \\
            &= log  \mathbb{E}_{z_{S:T} \sim P_{infr}} \dfrac{P(X_{S:T}|z_{S:T}) P_{pst}(z_{S:T})}{P_{infr}(z_{S:T})} \\
            &\geqslant    \mathbb{E}_{z_{S:T} \sim P_{infr}} log \frac{P(X_{S:T}|z_{S:T}) P_{pst}(z_{S:T})}{P_{infr}(z_{S:T})} \\
            &= \mathbb{E}_{z_{S:T} \sim P_{infr}} log P(X_{S:T}|z_{S:T}) \nonumber \\
            &\qquad\qquad - \mathbb{E}_{z_{S:T} \sim P_{infr}} log \frac{P_{infr}(z_{S:T})}{P_{pst}(z_{S:T})} \\ 
            &= \mathbb{E}_{z_{S:T} \sim P_{infr}} log \prod_{t} P(X_t|X_{S:t-1}, z_{S:t})  \nonumber \\
            &\qquad\qquad - \int_{z_{S:T}} P_{infr}(z_{S:T}) log \frac{P_{infr}(z_{S:T})}{P_{pst}(z_{S:T})} \\
            &= \sum_{t=S}^{T} \mathbb{E}_{z_t \sim P_{infr}} log P(X_t|X_{S:t-1},z_{S:t}) \nonumber \\
            &\qquad\qquad - \int_{z_{S:T}} P_{infr}(z_{S:T}) log \frac{P_{infr}(z_{S:T})}{P_{pst}(z_{S:T})} \label{eql:exclusive_half}
        \end{align}
    }
    The second term of \eqref{eql:exclusive_half} can be further derived as:
    {\small
        \begin{align}
            & \int_{z_{S:T}} P_{infr}(z_{S:T}) \cdot log \frac{P_{infr}(z_{S:T})}{P_{pst}(z_{S:T})} \nonumber \\
            &= \int_{z_S^{infr}}\int_{z_{S+1}^{infr}}...\int_{z_{T}^{infr}} \Big[\prod_{t=S}^T P_{infr}(z_t) \cdot \sum_t log \frac{P_{infr}(z_t)}{P_{pst}(z_t)}\Big] \\
            &= \sum_{t=S}^{T} \mathbb{E}_{z_t \sim P_{infr}} log \frac{P_{infr}(z_t)}{P_{pst}(z_t)} \\
            &= \sum_{t=S}^{T} D_{KL}\big(P_{infr}(z_t) || P_{pst}(z_t)\big) \label{eql:kl_inference1}
        \end{align}
    }
    Put \eqref{eql:kl_inference1} into \eqref{eql:exclusive_half}, we have:
    {\small
      \begin{align}
          & log P(X_{S:T}) \geqslant \sum_{t=s}^{T} \Big[ \mathbb{E}_{z_t \sim P_{infr}} log P(X_t|X_{S:t-1},z_{S:t}) \nonumber \\
          & - D_{KL}\Big(P_{infr}(z_t) || P_{pst}(z_t)\Big) \Big] \label{eq3}
      \end{align}
    }
    In our full loss, reducing the pixel prediction loss $L_1(X_t, \tilde{X}_t^{infr})$ maximizes the $\mathbb{E}_{z_t \sim P_{infr}} log P(X_t|X_{S:t-1},z_{S:t})$ in \eqref{eq3}. Similarly, reducing the exclusive KL divergent loss $D_{KL}\Big(P_{infr}(z_t) || P_{pst}(z_t)\Big)$ would maximizes $- D_{KL}\Big(P_{infr}(z_t) || P_{pst}(z_t)\Big)$ in \eqref{eq3}.

\section{Theoretical Explanation of Pixel Reconstruction Loss and Inclusive KL loss}
\label{appendix:B}
    Because the neural network has the tendency to use the easiest way to fit the objective, if we only maximize the right side of \eqref{eq3}, \(LSTM_{pst}\) will ignore the information introduced by \(X_t\). The \(P_{pst}(z_t)\) and \(P_{infr}(z_t)\) will degenerate to a convenient fixed value to lower the \(D_{KL}\). 
    To prevent the degeneration, we also need to maximize a sequence's probability \(P(X_{S:T})\) from the Posterior Module. The logarithm of this probability has the following lower bound:
    {\small
        \begin{align}
            & log P(X_{S:T})  \nonumber \\
            &= log \int_{z_{S:T}} P(X_{S+1:T}|z_{S:T})P_{infr}(z_{S:T})\frac{P_{pst}(z_{S:T})}{P_{pst}(z_{S:T})} \\
            &= log  \mathbb{E}_{z_{S:T} \sim P_{pst}} \dfrac{P(X_{S:T}|z_{S:T}) P_{infr}(z_{S:T})}{P_{pst}(z_{S:T})} \\
            &\geqslant    \mathbb{E}_{z_{S:T} \sim P_{pst}} log \frac{P(X_{S:T}|z_{S:T}) P_{infr}(z_{S:T})}{P_{pst}(z_{S:T})} \\
            &= \mathbb{E}_{z_{S:T} \sim P_{pst}} log P(X_{S:T}|z_{S:T}) \nonumber \\
            &\qquad\qquad - \mathbb{E}_{z_{S:T} \sim P_{pst}} log \frac{P_{pst}(z_{S:T})}{P_{infr}(z_{S:T})} \\ 
            &= \mathbb{E}_{z_{S:T} \sim P_{pst}} log \prod_{t} P(X_t|X_{S:t-1}, z_{S:t})  \nonumber \\
            &\qquad\qquad - \int_{z_{S:T}} P_{pst}(z_{S:T}) log \frac{P_{pst}(z_{S:T})}{P_{infr}(z_{S:T})} \\
            &= \sum_{t=S}^{T} \mathbb{E}_{z_t \sim P_{pst}} log P(X_t|X_{S:t-1},z_{S:t}) \nonumber \\
            &\qquad\qquad - \int_{z_{S:T}} P_{pst}(z_{S:T}) log \frac{P_{pst}(z_{S:T})}{P_{infr}(z_{S:T})} \label{eql:inclusive_half}
        \end{align}
    }
    The second term of \eqref{eql:inclusive_half} can be further derived as:
    {\small
        \begin{align}
            & \int_{z_{S:T}} P_{pst}(z_{S:T}) \cdot log \frac{P_{pst}(z_{S:T})}{P_{infr}(z_{S:T})} \nonumber \\
            &= \int_{z_S^{pst}}\int_{z_{S+1}^{pst}}...\int_{z_{T}^{pst}} \Big[\prod_{t=S}^T P_{pst}(z_t) \cdot \sum_t log \frac{P_{pst}(z_t)}{P_{infr}(z_t)}\Big] \\
            &= \sum_{t=S}^{T} \mathbb{E}_{z_t \sim P_{pst}} log \frac{P_{pst}(z_t)}{P_{infr}(z_t)} \\
            &= \sum_{t=S}^{T} D_{KL}\big(P_{pst}(z_t) || P_{infr}(z_t)\big) \label{eql:kl_inference2}
        \end{align}
    }
    Put \eqref{eql:kl_inference2} into \eqref{eql:inclusive_half}, we have:
    {\small
      \begin{align}
          & log P(X_{S:T}) \geqslant \sum_{t=s}^{T} \Big[ \mathbb{E}_{z_t \sim P_{pst}} log P(X_t|X_{S:t-1},z_{S:t}) \nonumber \\
          & - D_{KL}\Big(P_{pst}(z_t) || P_{infr}(z_t)\Big) \Big] \label{eq4}
      \end{align}
    }
    In our full loss, reducing the pixel reconstruction loss $L_1(X_t, \tilde{X}_t^{pst})$ maximizes the $\mathbb{E}_{z_t \sim P_{pst}} log P(X_t|X_{S:t-1},z_{S:t})$ in \eqref{eq4}. Similarly, reducing the inclusive KL divergent loss $D_{KL}\Big(P_{pst}(z_t) || P_{infr}(z_t)\Big)$ would maximizes $- D_{KL}\Big(P_{pst}(z_t) || P_{infr}(z_t)\Big)$ in \eqref{eq4}.

\section{Architecture and Training Details}
    \label{appendix:C}
    \(Encoder\), \(Extractor\) and \(Decoder\) use the same architecture of DCGAN. For step \(S\) and \(T\), feature maps of all layers in \(Encoder\) will be gathered as a multi-scale residuals \(ctn_{S}\) and \(ctn_{S}\) to help reconstruct the static content. \(LSTM_{infr}\), \(LSTM_{pst}\) and \(LSTM_{dyn}\) use the structure of one layer ConvLSTM. The output dimensions of our modules are listed in Table \ref{tab:dimension}. Our reported result is created using one extended references frame on each side (totally 2 extended reference frames). These two extra references bring extra long-term information into Reference module. However, frames too far away from the interval would contain too much unrelated information. We also tested on 4 extended reference frames and find the benefit is insignificant. 
    
    For the evaluation presented in our paper, all datasets have been trained with a input frame dimension of \(64 \times 64\) with a interval of 7 frames. We also train on KTH with a input frame dimension of \(128 \times 128\) (See Section 2 in the video web page) and BAIR with intervals of 9 frames (See Section 1 in the video web page). We use standard Adam optimizer with 0.5 as the first momentum decay rate. All settings of the hyper parameters for different datasets are shown in \ref{tab:hyper}. The \(\beta\) is initially set to 1 and gradually reduce to 0.4. To prevent the accumulation of errors, we will first roll back the cell state of \(LSTM_{infr}\) to \(t-1\), and then input \(h_t\) after inferring \(N_{infr}(\mu_t,\sigma_t)\). This operation has been proved to be crucial to our result. On the early stage of the training, a less meaningful \(\hat{h}_t\) would accumulatively disturb the cell state of \(LSTM_{infr}\) and lead to a slow convergence.  
    \begin{table}[!t]
    \begin{tabular}{|c|c|c|c|}
    Feature & Dim 1 & Dim 2 & Dim 3 \\
    \(C_{start}\)        & 4                   & 4                  & 256        \\
    \(C_{end}\)         & 4                   & 4                  & 256        \\
    \(h_{t}\)         & 4                   & 4                  & 256        \\
    \(\hat{h}_t\)         & 4                   & 4                  & 256        \\
    \(\sigma_t\) SMMNIST \& KTH         & 4                   & 4                  & 32         \\
    \(\mu_t\) SMMNIST \& KTH       & 4                   & 4                  & 32         \\
    \(\sigma_t\) BAIR       & 4                   & 4                  & 64         \\
    \(\mu_t\) BAIR        & 4                   & 4                  & 64        
    \end{tabular}
    \caption{The dimensionalities of different features}
    \label{tab:dimension}
    \end{table}
    \begin{table}[!t]
    \begin{tabular}{|c|c|c|c|}
    Training Parameters & SMMNIST  & BAIR     & KTH      \\
    \(\alpha\)                    & 0.002    & 0.0002   & 0.0002   \\
    \(\beta\)                    & 1 to 0.4 & 1 to 0.4 & 1 to 0.4 \\
    Map Weight \(\eta\)                   & N/A & 2 & 3
    \end{tabular}
    \caption{Hyper parameters for training on different datasets}
    \label{tab:hyper}
    \end{table}
\section{Dataset Details}
    \label{appendix:D}
    SMMNIST: Sequences were generated on the fly by randomly choosing two digits from MNIST: 50k digits from MNIST training set for training, 10k digits for validation, and 10k in MNIST testing set for testing.
    We create the ground truth video frames as 16 fps.
    
    KTH: We used person 1-14 (1337vids) for training, 15-16(190vids) for validation and 17-25 (863vids) for testing. We sample the ground truth video frames as 12 fps.
    
    Bair: By default, we use 40000 scenes for training, 3863 for validation and 256 for testing. We sample the ground truth video frames as 16 fps.
    
    UCF101: The dataset contains 101 realistic human actions taken in a wild and exhibits various challenges, such as background clutter, occlusion, and complicated motion. The training set contains 3223 video sequences with varying length, and the test set contains 557 video sequences. We sample the video as 16 fps so that the input reference frames for the network are 2 fps. 
\section{More Results}
    \label{appendix:E}
\begin{figure}[!htb]
    \centering
      \includegraphics[width=\linewidth]{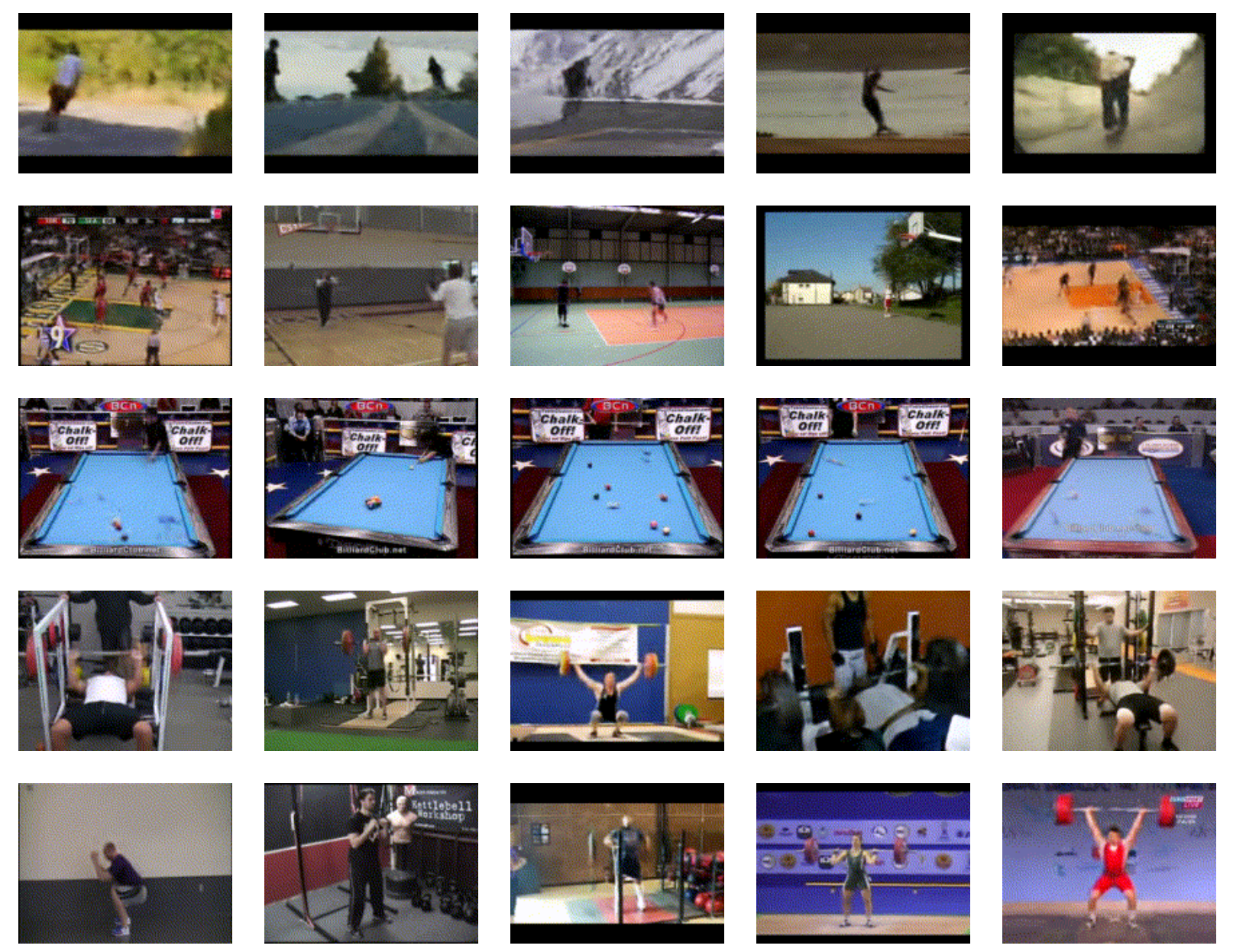}
      \caption{a snapshot of the gifs in the ``video\_result.html''}
      \label{fig:teaser}
    \end{figure}

\begin{figure*}[!htb]
    \centering
      \includegraphics[width=\textwidth]{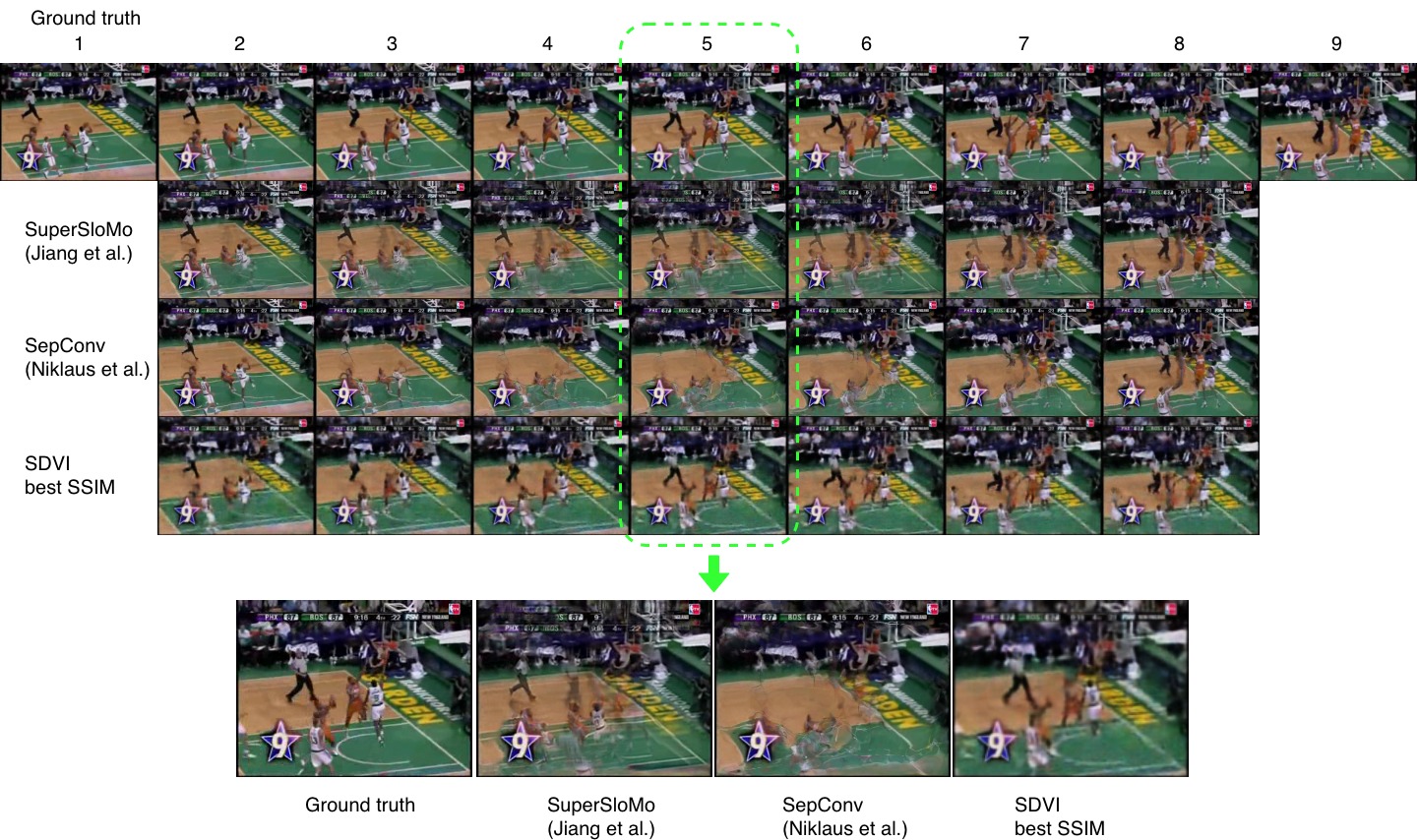}
      \caption{A more complicated UCF101 example: a real basketball video sequence involving multiple objects. Our method can model the dynamic correctly and generate better moving objects than SuperSloMo and SepConv.}
      \label{fig:stochastic_UCF_dunk}
    \end{figure*}

    Figure \ref{fig:stochastic_UCF_dunk} provides another UCF101 comparision. Results of more conditions and more promising results can be found in the ``video\_result.html''. Please open the webpage in your browser to see (the web page contains gif videos such as: Figure \ref{fig:teaser}).  We use the full datasets and train all actions together. We will include more videos generated by our ablation studies and the comparative models in our project website.
    
    \begin{figure}[!htb]
    \centering
      \includegraphics[width=\linewidth]{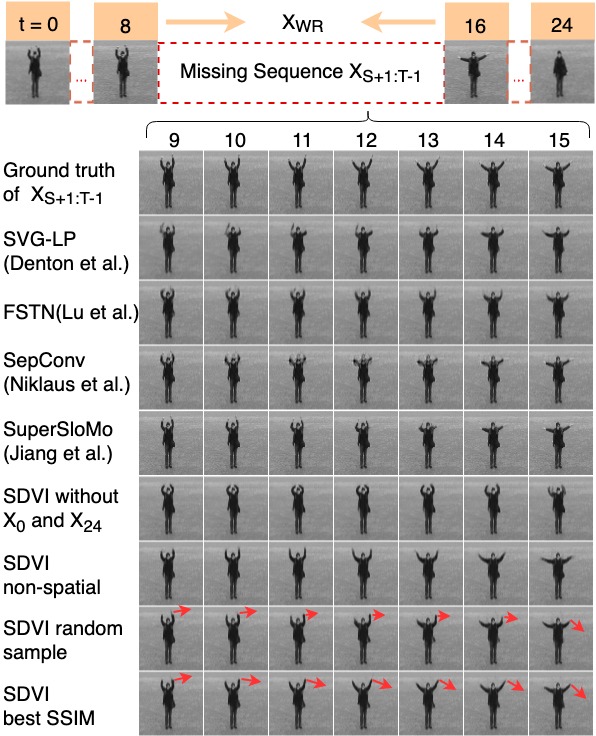}
      \caption{ shows the full comparisons for he wave action.}
      \label{fig:stochastic_KTH2_full}
    \end{figure}

\end{appendices}
\end{document}